\documentclass{article}
\usepackage{graphicx}
\usepackage{booktabs}
\usepackage{wrapfig}
\usepackage{subcaption}
\usepackage{multirow}
\usepackage{float}
\usepackage{subcaption}
\usepackage{algorithm}
\usepackage{algpseudocode}
\usepackage[preprint]{corl_2026} 

\title{SynthICL: Scalable In-context Imitation Learning with Synthetic Data}

\author{
  Cheng Qian\\
  The Robot Learning Lab \\
  Imperial College London\\
  \And
  Ruomeng Fan \\
  The Robot Learning Lab \\
  Imperial College London\\
  \And
  Yifei Ren \\
  The Robot Learning Lab \\
  Imperial College London\\
  \And
  Yilong Wang \\
  The Robot Learning Lab \\
  Imperial College London\\
  \And
  Edward Johns \\
  The Robot Learning Lab \\
  Imperial College London\\
}

\begin{document}
\maketitle


\begin{abstract}
In-context imitation learning (ICIL) enables robots to learn new tasks from a small number of demonstrations by conditioning a pre-trained policy on task-specific examples, without retraining at test time. Despite this promise, training generalizable and scalable in-context imitation policies remains an open challenge. We present SynthICL, a scalable framework that trains ICIL policies entirely from RGB-only synthetic data. Specifically, we build a data generation pipeline to produce high-fidelity ICIL data and train a flow-matching transformer policy on the resulting dataset. SynthICL avoids the need for depth sensing, precise camera calibration, and real-world training data in prior approaches, offering a simpler and more scalable alternative. We further incorporate subgoal prediction by training the model to predict the next subgoal images, enabling more precise and visually grounded control. Evaluated on 16 unseen real-world manipulation tasks, SynthICL achieves an average success rate of 79\% with only one demonstration provided at test time and outperforms prior methods. Project page: \url{https://synth-icl.github.io}


\end{abstract}


\begin{figure}[htbp]
    \centering
    \includegraphics[width=\linewidth]{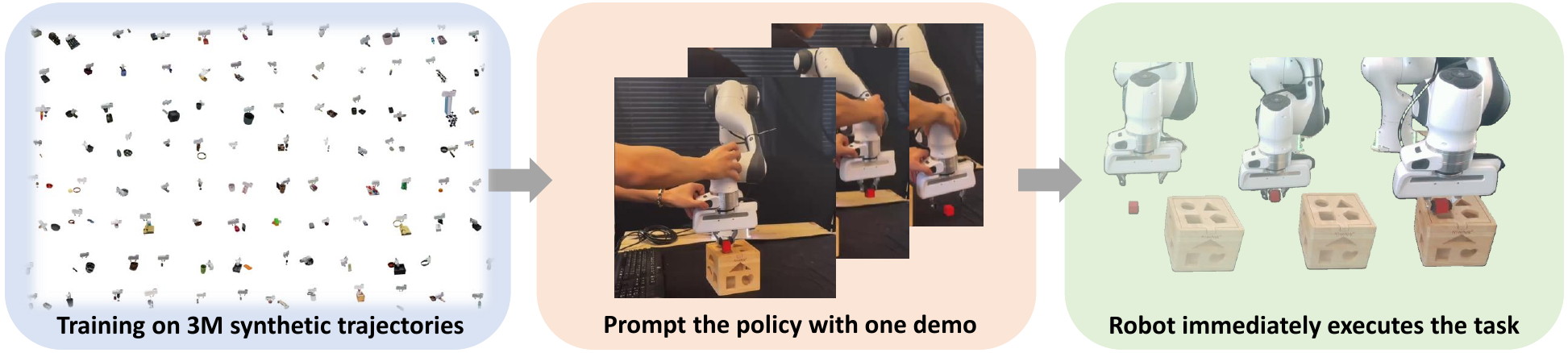}
    \caption{SynthICL is a framework that combines a pipeline for generating high-fidelity data for in-context learning, with a flow-matching transformer policy that predicts actions conditioned on contextual demonstrations. Once trained, the model can be directly deployed to unseen real-world tasks with only one demonstration provided at test time, with instant generalisation to novel object poses.}
    \label{fig:overview}
\end{figure}
\section{Introduction}
In-context imitation learning (ICIL) \citep{vosylius2024instant, fu2025icrt, di2024keypoint} has recently emerged as a promising paradigm for test-time robot learning. In contrast to conventional behavioral cloning methods \citep{pi0_2024, groot2024, kim2024openvla, team2024octo}, which typically rely on large datasets of task-specific demonstration data and fine-tuning, ICIL enables a single policy to infer the task from one or a small number of demonstrations provided at test time, then execute the desired behavior instantly without any weight updates to the policy. 

However, training ICIL policies to be capable of test-time generalisation from just one demonstration, requires large amounts of diverse data. Existing ICIL methods \citep{fu2025icrt, nguyen2026iclr} typically rely on large-scale real-world data collection, making them costly to scale. Alternatively, \citet{vosylius2024instant} proposes a data generation scheme that generates synthetic demonstrations for ICIL training. However, their method relies on point-cloud observations, which introduce two key limitations. First, point clouds suffer from a substantial sim-to-real gap: real-world point clouds are often noisy, incomplete, and dependent on accurate depth sensing and camera calibration, whereas simulated point clouds are typically clean and idealized. As a result, it requires real-world co-training to bridge this gap, reducing its scalability. Second, it uses pure point-cloud representation, discarding rich RGB appearance information, limiting the policy's ability to exploit fine-grained visual cues and making it less effective in tasks where objects have similar geometry but differ in appearances.

This motivates a question: can RGB-only synthetic training directly produce generalizable in-context policies? To answer this question, we propose SynthICL (see Figure~\ref{fig:overview}). Compared to existing approaches, SynthICL eliminates the need for depth sensing, precise camera calibration, and real-world training data. To support sim-to-real transfer, we implement a data generation pipeline in Isaac Sim \citep{NVIDIA_Isaac_Sim} that produces high-quality synthetic context–task trajectories and organizes them into aligned context-task pairs. We then train a flow-matching transformer policy \citep{lipman2022flow, liu2022rectified} on the resulting dataset. Since SynthICL is trained entirely from synthetic trajectories, subgoal states are available without additional annotation. We exploit this property by introducing an auxiliary subgoal prediction head that predicts the next subgoal image during training. This auxiliary objective encourages the model to reason jointly over the context and the current task state, and learn visually grounded subgoal representations. Our experiments further show that this subgoal prediction objective is critical for improving policy robustness and overall performance.


Our contributions are as follows: (1) we propose SynthICL, a scalable framework for training generalizable RGB-based in-context imitation learning policies entirely from high-fidelity synthetic data; (2) we introduce a novel in-context subgoal prediction objective that encourages visually grounded subgoal reasoning and substantially improves policy robustness and performance; and (3) we show through extensive simulation and real-world experiments that SynthICL outperforms prior methods on diverse manipulation tasks without any real-world training data and depth sensing.

\section{Related Works}
\label{sec:related works}
\paragraph{Imitation Learning For Robot Manipulation.}
Imitation learning \citep{hussein2017imitation, chi2025diffusion, zhao2023learning} has been widely used in robot manipulation to learn policies from expert demonstrations. Recent Vision-Language-Action models (VLAs) \citep{team2024octo, pi0_2024, groot2024} improve generalization by training on large-scale multi-task datasets and conditioning policy behavior on natural language instructions. However, these approaches remain limited when language is ambiguous or insufficient for specifying the desired behavior, and they often require additional fine-tuning to adapt to new tasks.

\paragraph{In-context Imitation Learning.}
In context learning has recently been introduced to robotics as a promising framework for enabling robots to adapt to unseen tasks from only a few demonstrations at test time \cite{fu2025icrt, vosylius2024instant, vosylius2023few, di2024keypoint, jain2024vid2robot, nguyen2026iclr, wang2025oneshotdualarmimitationlearning}. Instant Policy \cite{vosylius2024instant} (IP) is one of the works most related to ours. It introduces pseudo-demonstrations as training data, which are generated in simulation by sampling object-centric waypoints and constructing semantically consistent trajectories under randomized scene configurations. Our method also leverages pseudo-demonstrations for scalable data generation, but uses RGB observations rather than point clouds. Another closely related work, ICRT \citep{fu2025icrt}, formulates policy learning as autoregressive next-token prediction conditioned on contextual demonstrations. While ICRT also uses RGB observations, it requires large amounts of real-world demonstration data for training, which is costly to collect.

\paragraph{Subgoal Prediction for Imitation Learning.}
Incorporating subgoals has been shown to be an effective way to help policies model task progress and condition actions on intermediate goal states \citep{nguyen2026iclr, sharma2019third, lee2019selective, pertsch2020keyin, wen2021keyframe, lynch2020learning, ni2024generate, kang2025taksie, hatch2024ghil}. \citet{nguyen2026iclr} incorporates visual trace reasoning into ICIL by augmenting the contextual demonstrations with visual traces. The visual traces is represented as key pixels on the image. At test time, the model predicts future visual traces together with actions to improve task execution. In contrast, SynthICL directly predicts subgoal images instead of sparse key pixels to encourage the policy representation to capture the visual correspondence between the context and the current task images. Furthermore, we do not use predicted subgoals as explicit inference-time guidance, preserving a simple end-to-end policy at inference time.

\section{Methods}

\label{sec:methods}
SynthICL learns an RGB-based ICIL policy entirely from a large-scale synthetic dataset and can adapt to novel real-world tasks from a single test-time demonstration. Given one demonstration of a task, the policy conditions on the visual context and the current RGB observation to predict action chunks. Our method has three key ingredients: (i) a data generation pipeline that produces high-fidelity context-task trajectory pairs, (ii) a context-conditioned policy architecture, and (iii) an auxiliary subgoal image prediction objective. We first define the ICIL problem setting, then describe our data generation pipeline and model architecture.

\subsection{Problem Formulation}
We formulate ICIL as follows. At test time, the policy is given a single context demonstration
\(C\) of a novel task, together with the current observation \(o_t\) in a new scene. The goal is to infer the action \(a_t\) that enables the robot to complete the task specified by the context demonstration. The context demonstration is represented as an observation trajectory
\(C=\{\tilde{o}_j\}_{j=1}^{T}\), where each observation \(\tilde{o}_j\) consists of segmented RGB images \(\{\tilde{I}_{j}^{v}\}_{v=1}^{V}\) captured from \(V\) camera views. Similarly, the current observation \(o_t\) consists of multi-view RGB images \(\{I_t^{v}\}_{v=1}^{V}\). We do not use proprioceptive states, as we find that excluding them improves generalization, consistent with prior observations in \citet{zhao2025you}. The action \(a_t\) contains an end-effector displacement and a gripper command.
The policy \(\pi_\theta\) is trained to model the conditional action distribution
\(p(a_{t:t+H-1} \mid o_t, C)\), where \(a_{t:t+H-1}\) denotes an action chunk over a prediction horizon \(H\).

\begin{figure}[htbp]
    \centering
    \includegraphics[width=\linewidth]{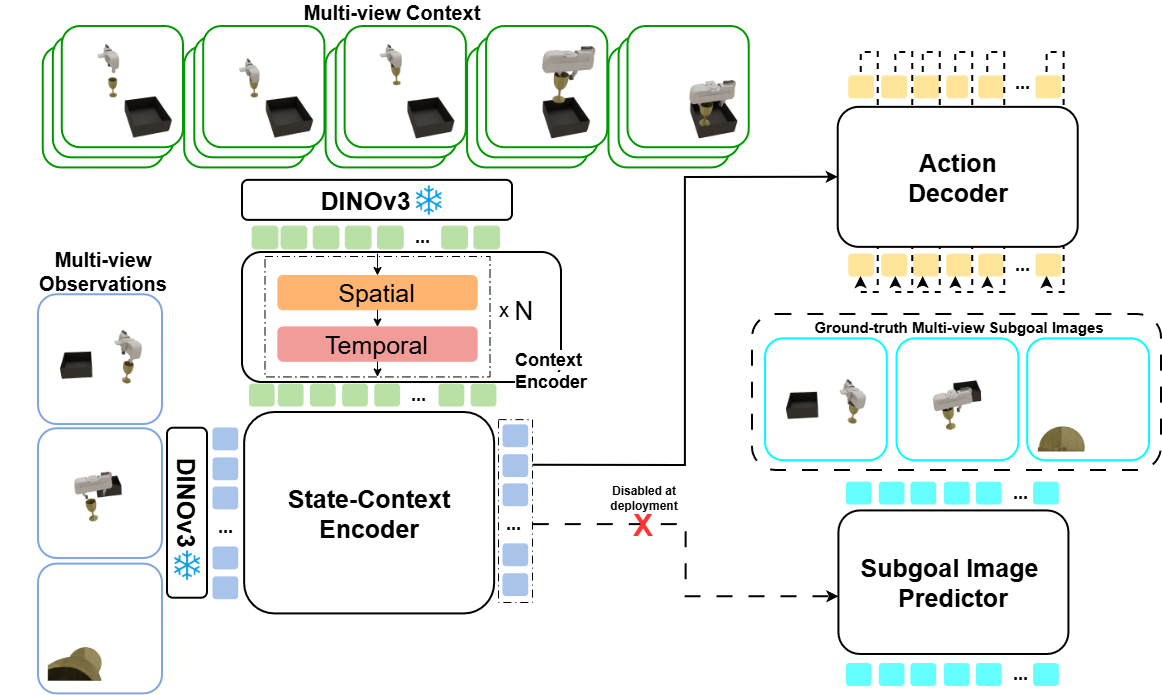}
    \caption{
    Overview of the proposed model architecture. The policy takes context demonstration and the current observation as input, encodes them through visual and temporal modules, and predicts robot actions for closed-loop control. In addition, it predicts subgoal multi-view images as an auxiliary training objective.
    }
    \label{fig:model_architecture}
\end{figure}
\subsection{Data Generation}
\label{data_generation}
To train SynthICL at scale, we generate synthetic context-task trajectory pairs rather than conventional simulation demonstrations, mainly following the \emph{pseudo-demonstrations} generation pipeline of Instant Policy (IP) \citep{vosylius2024instant}. The key idea of pseudo demonstrations is rendering semantically consistent context–task pairs, but without physics simulation. For each episode, we first sample one or a few objects from RoboTwin-OD~\citep{mu2024robotwin, chen2025robotwin} and ShapeNet~\citep{chang2015shapenet} datasets, apply random texture from DTD dataset~\citep{cimpoi14describing}, then sample object-centric waypoints around these objects, including gripper grasping and releasing states. The gripper is then moved through the sampled waypoints to form trajectories that resemble manipulation behaviors. By randomizing objects and waypoints configurations, diverse trajectories corresponding to the same semantic task can be produced. \footnote{Here, trajectories are considered semantically consistent if the relative poses between the robot end-effector and each object are kept consistent at corresponding waypoints. For grasping interactions, when the gripper reaches a grasping waypoint, the target object is attached to the gripper and moves with it until being released.} 

In addition to this pseudo demonstrations generation strategy from IP, we improve the pipeline with two important modifications. First, instead of using PyBullet~\citep{coumans2021}, as in IP, we use Isaac Sim~\citep{NVIDIA_Isaac_Sim} to render substantially higher-fidelity RGB images and reduce the sim-to-real visual gap. Second, the original pipeline randomly samples grasp poses, which can produce many invalid or unnatural grasp poses, reducing data quality and potentially degrading policy learning. To address this, we pre-compute multiple valid grasp poses for each object using GraspNet~\citep{fang2020graspnet, fang2023robust}. During data generation, for the waypoints that involve grasping, we randomly sample one of these pre-computed grasp poses. 

\subsection{Model Architecture}
Our model consists of four main components: a context encoder, a state-context encoder, an action decoder, and an auxiliary subgoal image decoder. The context encoder encodes context demonstration into a compact spatio-temporal representation. The state-context encoder then retrieves features from the encoded context that are relevant to the current observation. Based on these extracted features, the action decoder predicts the robot action. In addition, we introduce a subgoal image decoder as an auxiliary training objective.

\paragraph{Context Encoder.}
The input to the context encoder is a sequence of observations from the context demonstration \(C\). Each observation contains segmented RGB images from multiple camera views. We encode each segmented RGB image using a frozen DINOv3 vision encoder and add temporal positional embeddings to distinguish frames at different time steps and learnable embeddings for different camera views. The resulting tokens are then passed into a spatio-temporal transformer with decomposed spatial and temporal attention blocks \citep{ma2024latte, bertasius2021space}. Spatial attention operates over patch tokens within each frame, while temporal attention operates across frames at corresponding patch locations. This avoids full attention over all space-time tokens while preserving spatial and temporal information.
The output of this transformer serves as the encoded context representation and is passed to the state-context encoder.

\paragraph{State-context Encoder.}
Given the encoded context representation, the state-context encoder extracts and encodes the information that is most relevant to the current task state. We encode the current segmented multi-view RGB images in the same way as in the context encoder with DINOv3. The resulting feature map serves as queries, while the encoded context features serve as keys and values, and are fed into a cross-attention transformer. 
The resulting task-relevant representation is then passed to the action decoder and subgoal image predictor.

\paragraph{Action Decoder.}
The action decoder predicts an action chunk conditioned on the encoded task-relevant representation and is formulated as a conditional flow-matching problem. Let \(x_1=a_{t:t+H-1}\) denote the ground-truth clean action chunk and let \(x_0 \sim \mathcal{N}(0,\mathbf{I})\) denote a noise sample with the same dimension. We construct an interpolated action chunk
\[
x_\tau = (1-\tau)x_0 + \tau x_1, \qquad \tau \sim \mathcal{U}(0,1),
\]
where \(\tau\) denotes the flow-matching time. The action decoder takes the task-relevant feature, the interpolated action chunk \(x_\tau\), and the time \(\tau\) as input, and predicts a velocity field \(v_\theta(x_\tau,\tau \mid z_t)\), where \(z_t\) is the extracted task-relevant representation.

The decoder is trained to match the target flow from noise to data:
\[
\mathcal{L}_{\mathrm{action}}
=
\left\|
v_\theta(x_\tau,\tau \mid z_t) - (x_1-x_0)
\right\|_2^2 .
\]
At inference time, we sample a Gaussian noise action chunk and integrate the learned velocity field from \(\tau=0\) to \(1\) using 10 denoising steps. The predicted chunks are executed with temporal ensembling for smooth control \citep{zhao2023learning}.

\paragraph{Subgoal Image Predictor.}
In pseudo-demonstrations, each trajectory is associated with a sequence of waypoints, where each waypoint can be viewed as a subgoal state. We define the next subgoal as the rendered multi-view observation at the next waypoint. We exploit this property by introducing an auxiliary subgoal image prediction head in parallel with the action decoder. Conditioned on the extracted task-relevant representation, this head predicts the multi-view next subgoal image. In this way, the model is encouraged to reason about task progress and visually grounded intermediate states, rather than merely regressing actions. The subgoal image prediction head is architecturally separate from the action decoder and does not share attention layers with it. This design allows the subgoal head to be disabled at test time, while improving the policy representation during training. In Section~4, we compare this design with another paradigm, in which subgoal prediction and action generation are performed within a shared model.

We train the subgoal image prediction head using an image reconstruction loss. Let \(\{I^{\mathrm{sub},v}_{t}\}_{v=1}^{V}\) denote the ground-truth multi-view images of the next subgoal state, and let \(\{\hat{I}^{\mathrm{sub},v}_{t}\}_{v=1}^{V}\) denote the corresponding images predicted by the subgoal head.. The subgoal prediction loss is defined as an \(MSE\) reconstruction objective over all camera views:
\[
\mathcal{L}_{\mathrm{subgoal}}
=
\frac{1}{V}
\sum_{v=1}^{V}
\left\|
\hat{I}^{\mathrm{sub},v}_{t}
-
I^{\mathrm{sub},v}_{t}
\right\|_2^2 .
\]

The final training objective combines the flow-matching action loss and the auxiliary subgoal reconstruction loss:
\[
\mathcal{L}
=
\mathcal{L}_{\mathrm{action}}
+
\lambda_{\mathrm{subgoal}}
\mathcal{L}_{\mathrm{subgoal}},
\]
where \(\lambda_{\mathrm{subgoal}}\) weights the auxiliary subgoal prediction objective. In our experiments, we set \(\lambda_{\mathrm{subgoal}}=1.0\).
\section{Experimental Results}
\label{sec:result}
In this section, we evaluate SynthICL in both simulation and real-world settings. Our experiments are designed to answer four questions: (1) how SynthICL compares with other ICIL baselines, (2) how the improved pseudo-demonstration generation pipeline contributes to policy performance, (3) how the proposed method scales with increasing training data, and (4) how different design choices affect performance.

\subsection{Baselines}
We compare our method with three baselines: Instant Policy (IP)~\citep{vosylius2024instant}, ICRT~\citep{fu2025icrt} and SynthICL w/o SP. IP is the most closely related method to ours and serves as the primary baseline for comparing point-cloud-based and RGB-based in-context policies. ICRT is an RGB-based ICIL method, but with an auto-regressive model architecture. In addition, we evaluate an ablated variant of our method, SynthICL w/o SP, where SP denotes subgoal prediction, to measure the contribution of the auxiliary subgoal image prediction objective. For IP, we directly evaluate the released checkpoint. For ICRT, we train a policy with their architecture on our synthetic dataset.

\subsection{Simulation Experiments}
We conduct simulation experiments in RLBench \citep{james2020rlbench}. Our evaluation consists of 12 manipulation tasks. All evaluation tasks are unseen in the pseudo-demonstration training set. All methods are provided with observations from a three-camera setup, including a front view, a side view, and a wrist-mounted camera. We collect 100K pseudo demonstrations in PyBullet for training. During evaluation, only one demonstration is provided. We do not use Isaac Sim here, because RLBench uses the same renderer as PyBullet. Besides, the released IP checkpoint was trained on dataset that includes demonstrations for some of the RLBench tasks~\citep{vosylius2024instant}. Since our policy does not use demonstrations from these tasks during training, we report, for fairness, the success rates of the variant trained without those tasks data from the original IP paper. For each task, we evaluate 100 episodes with a maximum rollout horizon of 100 timesteps, and report the task success rate.

\begin{table*}[t]
\centering
\fontsize{7pt}{8pt}\selectfont
\setlength{\tabcolsep}{6pt}
\caption{RLBench task success rates with 100 rollouts each.}
\label{tab:sim_world_success}
\begin{tabular}{lcccc lcccc}
\toprule
\textbf{Task} & \textbf{IP} & \textbf{ICRT} & \textbf{Ours w/o SP} & \textbf{Ours}
&
\textbf{Task} & \textbf{IP} & \textbf{ICRT} & \textbf{Ours w/o SP} & \textbf{Ours} \\
\midrule
Rubbish in Bin        & 0.97 & 0.87 & 0.95 & 0.96 &
Cube in Shape Sorter  & 0.78 & 0.43 & 0.67 & 0.75 \\

Phone on Base         & 0.98 & 0.83 & 0.90 & 0.92 &
Press Button          & 0.60 & 0.48 & 0.65 & 0.68 \\

Meat on Grill         & 0.78 & 0.70 & 0.76 & 0.80 &
Lamp On               & 0.42 & 0.29 & 0.34 & 0.37 \\

Meat off Grill        & 0.77 & 0.64 & 0.74 & 0.78 &
Take Umbrella Out     & 0.88 & 0.72 & 0.80 & 0.81 \\

Basketball in Hoop    & 0.66 & 0.63 & 0.87 & 0.89 &
Slide Block to Target & 0.75 & 0.68 & 0.72 & 0.73 \\

Stack Cups            & 0.45 & 0.41 & 0.57 & 0.68 &
Water Plants          & 0.65 & 0.53 & 0.69 & 0.73 \\
\midrule
\textbf{Average} & 72.4\% & 60.1\% & 72.2\% & \textbf{75.0\%} &
 & & & & \\
\bottomrule
\end{tabular}
\end{table*}
\vspace{-10pt}
\paragraph{Results.} Table~\ref{tab:sim_world_success} reports the success rates of all methods on the RLBench tasks. Overall, SynthICL achieves the highest average success rate across the benchmark, indicating that large-scale synthetic data training enables strong generalization for RGB-based ICIL policies on novel tasks. Compared to point-cloud-based baseline (IP), our policy performs substantially better on tasks requiring appearance-based grounding. For example, in \textit{Stack Cups}, where the two cups share the same geometry but differ in color, our policy are better able to disambiguate the target object from visual appearance. Compared with SynthICL w/o SP, the full SynthICL model achieves higher average performance, especially on tasks that require higher precision, e.g., \textit{Cube in Shape Sorter}. This suggests that subgoal image prediction provides a useful auxiliary training signal, encouraging the policy to learn visually grounded subgoal representations for more accurate closed-loop control. 

\subsection{Real-world Experiments}
\begin{figure}[t]
    \centering
    \includegraphics[width=1.0\linewidth]{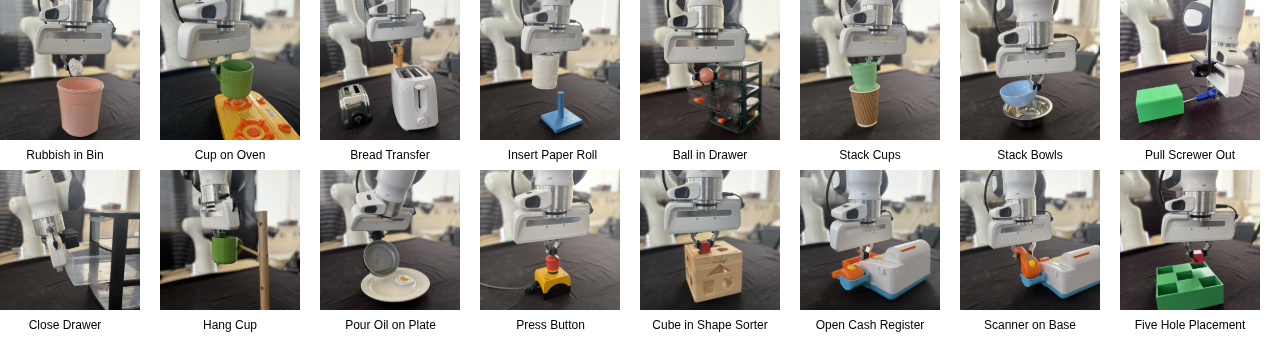}
    \caption{The 16 tasks we use in our real-world evaluation.}
    \label{fig:realbenchmarks}
\end{figure}

 We evaluate our method on 16 real-world manipulation tasks that cover a diverse range of manipulation skills (See Figure~\ref{fig:realbenchmarks}). The detailed description of each task can be found in Section~\ref{sec:real_world_task_descriptions}. We conduct real-world experiments using a Franka Research 3 robot with three RGB cameras: one wrist-mounted camera and two external cameras to provide front and side views. We obtain the segmentation using SAM~\citep{kirillov2023segment} and CUTIE~\citep{cheng2024putting}. For real-world experiments, we collect a dataset consisting of 3M trajectories in Isaac Sim and train the policy.

\begin{table*}[t]
\centering
\fontsize{7pt}{8pt}\selectfont
\setlength{\tabcolsep}{5pt}
\caption{Real-world task success rates with 20 rollouts each.}
\label{tab:real_world_success}
\begin{tabular}{lcccc lcccc}
\toprule
\textbf{Task} & \textbf{IP} & \textbf{ICRT} & \textbf{Ours w/o SP} & \textbf{Ours}
&
\textbf{Task} & \textbf{IP} & \textbf{ICRT} & \textbf{Ours w/o SP} & \textbf{Ours} \\
\midrule
Rubbish in Bin       & 18/20 & 13/20 & 16/20 & 17/20 &
Close Drawer         & 18/20 & 15/20 & 16/20 & 18/20 \\

Mug on Stove          & 16/20 & 8/20 & 16/20 & 19/20 &
Hang Cup             & 15/20 & 6/20 & 10/20 & 13/20 \\

Bread Transfer       & 15/20 & 11/20 & 14/20 & 17/20 &
Pour Oil on Plate    & 17/20 & 12/20 & 14/20 & 16/20 \\

Insert Paper Roll    & 16/20 & 10/20 & 12/20 & 15/20 &
Press Button         & 17/20 & 11/20 & 13/20 & 17/20 \\

Ball in Drawer       & 16/20 & 7/20 & 18/20 & 19/20 &
Cube in Shape Sorter & 13/20 & 5/20 & 10/20 & 15/20 \\

Stack Cups           & 10/20 & 12/20 & 17/20 & 18/20 &
Open Cash Register   & 15/20 & 2/20 & 5/20   & 11/20 \\

Stack Bowls          & 9/20 & 15/20 & 20/20 & 20/20 &
Scanner on Base      & 7/20 & 0/20 & 3/20  & 7/20 \\

Pull Screwdriver out & 18/20 & 12/20 & 16/20 & 18/20 &
Five Holes Placement & 12/20 & 5/20 & 9/20  & 13/20  \\
\midrule
\textbf{Average}     & 72.5\% & 45.0\% & 65.3\% & \textbf{79.1\%} &
                      &    &    &    &    \\
\bottomrule
\end{tabular}
\end{table*}
\vspace{-10pt}
\paragraph{Results.} Table~\ref{tab:real_world_success} reports the success rates of all methods on the 16 real-world manipulation tasks. Overall, SynthICL achieves the highest average success rate across the benchmark, showing that an RGB-based in-context policy trained entirely on synthetic pseudo-demonstrations can transfer effectively to diverse real-world manipulation tasks. Consistent with the simulation results, SynthICL performs substantially better on tasks that require distinguishing objects with similar geometry but different textures, such as \textit{Stack Bowls} and \textit{Stack Cups}, compared to the point-cloud-based baseline (IP). For tasks that require accurate spatial reasoning, such as \textit{Cube in Shape Sorter}, \textit{Scanner on Base}, and \textit{Five Holes Placement}, SynthICL still achieves the best performance despite using neither depth sensing nor real-world co-training data. This suggests that RGB-only synthetic training can support accurate real-world spatial reasoning, while point-cloud-based methods may suffer from noisy real-world depth observations and a larger point-cloud sim-to-real gap. However, on these tasks, SynthICL w/o SP achieves substantially lower success rates than the full SynthICL model. The performance gap between SynthICL and SynthICL w/o SP is larger in the real world (13\%) than in simulation (3\%), suggesting that subgoal image prediction is particularly effective in real-world tasks. ICRT achieves relatively lower success rates overall. We hypothesize that its autoregressive architecture is less suitable for policies trained only on clean synthetic data, as prediction errors may accumulate over long-horizon real-world execution.

\subsection{Data Generation}
\label{sec:data_generation}
We further compare policies training using the original pseudo-demonstration generation pipeline from IP~\citep{vosylius2024instant} with our improved pipeline, in order to evaluate the impact of the two modifications described in Section~\ref{data_generation}: higher-fidelity rendering and grasp-pose sampling. We generate datasets using these different synthetic data generation pipelines, each containing 3M trajectories. We then train the full SynthICL model on each dataset and evaluate the policies in real-world experiments to compare their performance.
\begin{wrapfigure}[13]{r}{0.35\textwidth}
    \centering
    \vspace{15pt}
    \includegraphics[width=0.33\textwidth]{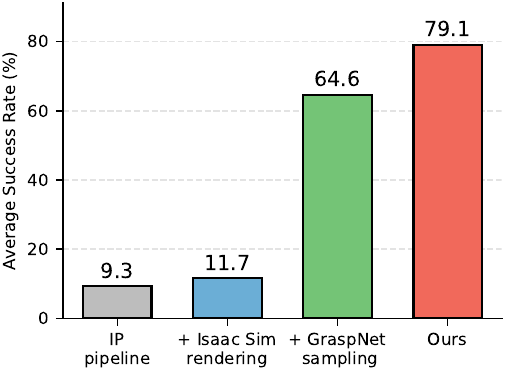}
    \caption{Comparison of SynthICL training with different data generation pipelines on real-world tasks.}
    \label{fig:data_generation_ablation}
    \vspace{-0pt}
\end{wrapfigure}
\paragraph{Results.} Figure~\ref{fig:data_generation_ablation} compares the average success rates of policies trained with different pseudo-demonstration generation pipelines. As shown in the figure, the pipelines without grasp pose sampling achieve only 9\% and 11\% success rates, respectively. This suggests that incorporating valid grasp pose sampling in data generation substantially improves the training data efficiency and results in better policy performance. 
We hypothesize that this improvement is because naively randomizing gripper poses can generate many uninformative or out-of-distribution grasp configurations, which may harm policy learning. In addition, without proper grasp pose sampling, the wrist-camera view can become largely uninformative because the target object may entirely occlude or fall outside the camera view.
Using Isaac Sim rendering instead of PyBullet improves the average success rate by 14\%, suggesting that using higher-fidelity RGB data for training can effectively enhance the sim-to-real capability of RGB-based ICIL policy. 

\begin{wrapfigure}[9]{r}{0.35\textwidth}
    \centering
    \vspace{-12pt}
    \includegraphics[width=0.35\textwidth]{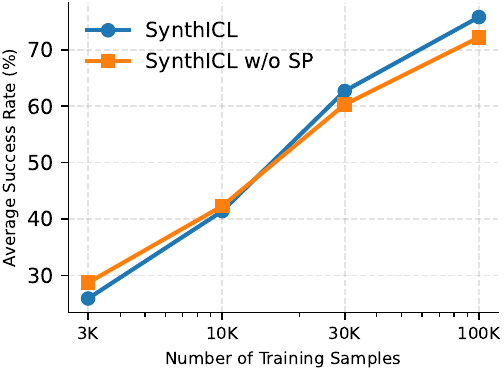}
    \caption{Data scaling results.}
    \label{fig:data_scaling}
\end{wrapfigure}
\subsection{Data Scaling}
We evaluate how policy performance changes with the scale of dataset. Specifically, we train SynthICL on datasets ranging from 3K to 100K trajectories, evaluate them on the simulation benchmark, and report the average success rate across all tasks.
\paragraph{Results.} As shown in Figure~\ref{fig:data_scaling}, both methods improve as the amount of synthetic training data increases, demonstrating that our framework can effectively benefit from increasing dataset scale. We further find that SynthICL w/o SP outperforms SynthICL when the dataset is small, suggesting that subgoal prediction benefits the policy only when sufficient training data are available. One possible explanation is that the auxiliary reconstruction objective requires enough diverse data to learn useful representations, and may otherwise compete with action learning in the low-data regime.
\subsection{Ablation Studies}
We further ablate two key design choices in SynthICL: the context encoding architecture and the subgoal prediction architecture. Specifically, we compare our decomposed spatial-temporal attention with standard full attention, and compare our separate action and subgoal prediction design with a joint prediction paradigm. We report the average success rate across all simulation tasks.
\begin{wraptable}[10]{r}{0.45\textwidth}
    \centering
    \vspace{13pt}
    \footnotesize
    \setlength{\tabcolsep}{4pt}
    \caption{Ablation of context encoding and subgoal prediction design choices.}
    \label{tab:model_design_ablation}
    \begin{tabular}{lcc}
    \toprule
    \textbf{Ablation} & \textbf{Variant} & \textbf{Avg. SR} \\
    \midrule
    \multirow{2}{*}{Context Encoder}
    & Full attention & 63.4\% \\
    & Spatial-temporal & 75.0\% \\
    \midrule
    \multirow{2}{*}{Subgoal Prediction}
    & Joint decoder & 76.1\% \\
    & Separate head & 75.0\% \\
    \bottomrule
    \end{tabular}
\end{wraptable}
\paragraph{Results.} Table~\ref{tab:model_design_ablation} summarizes the ablation results. The decomposed spatial-temporal context encoder outperforms standard full attention (+11.6\%), suggesting that separately modeling spatial and temporal interactions provides a more effective representation for multi-view context demonstration. For subgoal prediction, the separate head achieves performance comparable to the joint prediction paradigm, while reducing inference-time computation. We therefore adopt the separate design.



\section{Conclusion}
\label{sec:conclusion}

We have presented SynthICL, a scalable framework for learning RGB-based in-context imitation policies entirely from synthetic data. By training on large-scale, high-fidelity RGB data, SynthICL can be directly deployed to unseen real-world tasks with only one demonstration provided at test time, without requiring depth sensing, precise camera calibration, or real-world data co-training. We further proposed a novel in-context subgoal image prediction objective to encourage the policy to learn visually grounded subgoal representations. Through simulation and real-world experiments, we showed that SynthICL outperforms prior ICIL methods across diverse manipulation tasks. Our results show that large-scale RGB synthetic training is a viable path toward scalable in-context imitation learning.
\paragraph{Limitations.}Despite the promising results, SynthICL still has several limitations. First, our policy relies on segmented RGB images. Second, SynthICL still struggles with tasks that require very high precision or rich contact interactions, such as tight insertion, alignment, or manipulation involving complex physical contacts. Third, our current evaluation focuses mainly on relatively short-horizon manipulation tasks. However, we believe these limitations can be addressed by further improving the pseudo-demonstration generation pipeline, such as incorporating more diverse background randomization, introducing physics-based simulation, and generating long-horizon demonstrations.


\clearpage
\acknowledgments{If a paper is accepted, the final camera-ready version will (and probably should) include acknowledgments. All acknowledgments go at the end of the paper, including thanks to reviewers who gave useful comments, to colleagues who contributed to the ideas, and to funding agencies and corporate sponsors that provided financial support.}


\bibliography{reference}  
\newpage
\appendix
\section{Pseudo-Demonstration Generation Details}
\begin{figure}[h]
    \centering
    \includegraphics[width=1\linewidth]{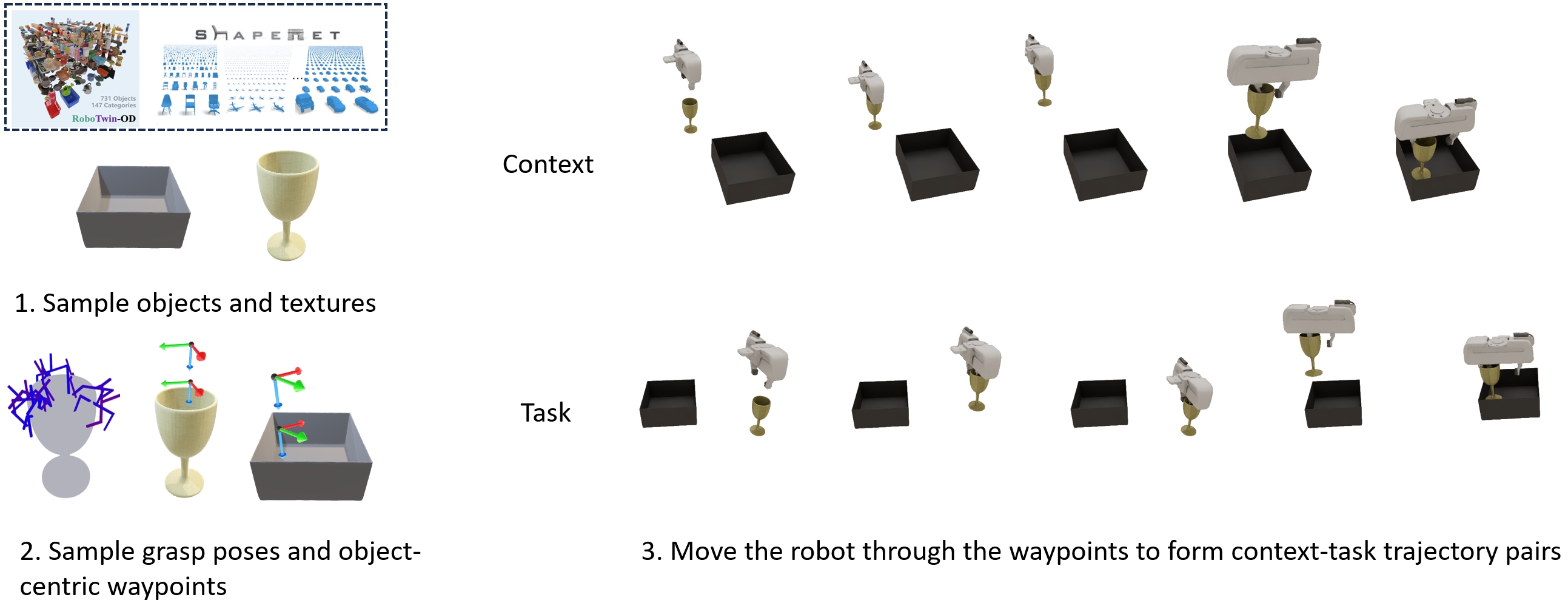}
    \caption{Overview of the pipeline of generating pseudo demonstrations.}
    \label{fig:datagenerationpipe}
\end{figure}

We generate each pseudo-demonstration in three steps (See Fig~\ref{fig:datagenerationpipe}). First, we sample one or two objects from RoboTwin-OD and ShapeNet~\citep{chang2015shapenet}. For ShapeNet, we keep only tabletop objects and remove large object categories such as chairs and tables. Second, we sample object-centric waypoints for each object, including grasping and releasing states. For each grasping state, we sample a grasp pose from precomputed GraspNet grasps. For grasp pose sampling, we use the full point-cloud of objects, and retain only grasp poses where the angle between the gripper and the ground plane is less than $45^\circ$ to simplify the data distribution. However, our method can be extended to all poses. Third, the object configuration is randomized, and the robot gripper is moved through these waypoints using linear interpolation to form context-task trajectory pairs: the object is attached to the gripper at grasping states and detached at releasing states. We choose 1 cm as the maximum interpolation step size. During rendering, we also apply slight randomization to the lighting conditions and camera poses in each episode to improve generalization. The context and task trajectories share the same waypoint sequence, while initial object poses, lighting, textures, and camera poses are randomized independently. Following Instant Policy~\citep{vosylius2024instant}, we further improve data efficiency by manually designing several common trajectory templates, including pick-and-place, pushing, and pouring, and oversampling them during training.

\section{Model Architecture Detail}
\begin{figure}[h]
    \centering
    \includegraphics[width=1\linewidth]{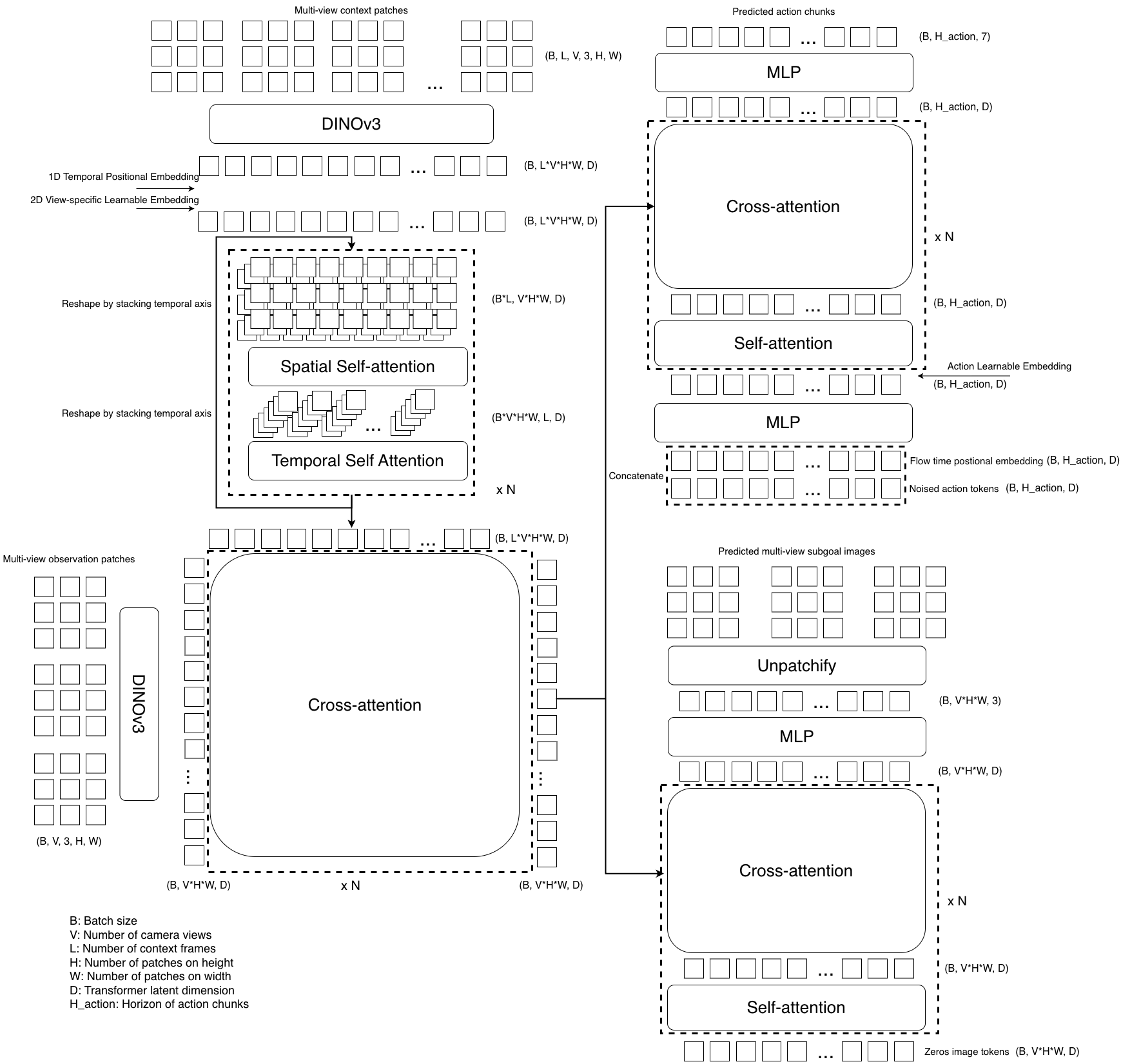}
    \caption{Details of the model architecture.}
    \label{fig:modeldetail}
\end{figure}

The detailed architecture of the SynthICL model is illustrated in Fig.~\ref{fig:modeldetail}.
\paragraph{Context Encoder.}
The context encoder encodes the visual observations from the context demonstration into a spatio-temporal representation. Each RGB image is first passed through a frozen DINOv3 backbone, and the resulting feature map is projected to the transformer hidden dimension with a \(1\times1\) convolution. We then flatten the feature map into patch tokens and add temporal positional embeddings as well as learnable camera-view embeddings. The tokens from all context frames and camera views are processed by a spatio-temporal transformer. Instead of applying full attention over all tokens, the encoder uses decomposed attention blocks: spatial attention models relations among patches within each frame, while temporal attention models relations among the same patches with different temporal indices. This allows the encoder to capture both object-level visual information and task progression in the demonstration. The output tokens are normalized and used as the context memory for the state-context encoder.

\paragraph{State-context Encoder.}
The state-context encoder aligns the current observation with the encoded context demonstration. We first encode the current multi-view RGB observation using the same frozen DINOv3 backbone and patch-tokenization process as the context encoder. These current-observation tokens are used as query tokens, while the context tokens produced by the context encoder are used as keys and values. The encoder is implemented as a cross-attention transformer. Through cross-attention, each current-observation token can retrieve relevant information from the demonstration, such as the corresponding object, target location, or task stage. The output is a task-relevant representation, which combines the current visual state with information from the context demonstration. This representation is then shared by the action decoder and the auxiliary subgoal image decoder.

\paragraph{Action Decoder.}
The action decoder follows an ACT-style \citep{zhao2023learning} transformer decoder but is adapted for conditional flow matching. Its input is a noisy action chunk. The action vector consists of the relative end-effector translation, relative rotation (represented in rotvec), and a binary gripper command. Each action vector is embedded with a linear projection. The scalar flow time is encoded with a sinusoidal timestep embedding and expanded across the action horizon. We concatenate the action embedding and the timestep embedding and pass them through an MLP, producing time-conditioned action tokens. These action tokens are used as decoder queries. A learnable horizon positional embedding is added to indicate the temporal index of each action in the predicted chunk. The decoder performs cross-attention from the action queries to the task-relevant representation. The decoder output contains one hidden token for each future action step. A final linear action head is applied independently to each token to predict the velocity of the corresponding action vector. During inference, the same decoder is repeatedly evaluated in an Euler solver to transform an initial Gaussian action chunk into the final predicted action chunk. 

\paragraph{Subgoal Image Predictor.}
The auxiliary subgoal image decoder is implemented as a transformer-based reconstruction head. It takes the task-relevant representation from the state-context encoder as conditioning and predicts the RGB images of the next subgoal state for all camera views. Unlike the action decoder, this branch directly reconstructs the target subgoal images. For each camera view, the decoder starts from an empty image-like input and divides it into non-overlapping patches. Each patch is projected into the transformer hidden dimension using a patch embedding layer. We add a fixed two-dimensional sinusoidal positional embedding to each patch token to encode its spatial location in the image. Since the decoder predicts multi-view subgoal observations, we also add a learnable view embedding to distinguish tokens from different camera views. The decoder is composed of multiple transformer decoder layers. Within each layer, the image tokens first perform self-attention, which allows information exchange across spatial patches and camera views. They then cross-attend to the task-relevant representation. After the transformer layers, the output tokens are normalized and passed through a patch-wise prediction head. The patch-wise prediction head is implemented as a small MLP/linear projection that maps each decoded patch token to the RGB pixel values of the corresponding image patch. The predicted patch values are then unpatchified to reconstruct full RGB images for all camera views. The decoder is trained with an MSE reconstruction loss between the predicted multi-view images and the ground-truth next-subgoal images.

\section{Implementation Detail}
\paragraph{Demo Preprocessing.}
During training, we use the rendered multi-view images at each waypoint, together with the initial observation, as context frames. Our context encoder can take up to six multi-view context frames (We use six context frames because this is sufficient for most short-horizon manipulation tasks. However, our model architecture can be extended to support more context frames with minor modifications.). If fewer than six frames are available, we pad the remaining entries with zeros. For RLBench experiments, since the demonstrations for each task are constructed from waypoints, we directly use the corresponding waypoint frames to prompt the policy. For real-world experiments, where demonstrations are recorded as continuous trajectories, we select context frames using a simple heuristic. We first include the initial frames and the frames at which the gripper state changes. We then include frames where the gripper velocity falls below a predefined threshold, as these frames typically correspond to changes in the gripper's motion direction or interaction phase. If the number of selected frames is less than six, we pad the remaining context with zeros. We only use visual observations from the context demonstration and do not provide the corresponding context actions to the policy. It makes the test-time demonstration easier to collect, since users only need to provide a visual trajectory rather than accurate action labels.

\paragraph{Data Augmentation.}
During data generation, we perturb each sampled waypoint with small pose noise to reduce overfitting. Specifically, we add translational noise of 0.5 cm and rotational noise of 5 degrees. This is important because wrist-camera observations can be nearly identical for repeated executions of the same waypoint. Without waypoint noise, the model may learn a shortcut: closing or releasing the gripper only when the current wrist-camera image closely matches the waypoint image in the context, rather than learning the underlying task structure. During training, to further improve generalization and robustness, we apply color jitter augmentation to all frames passed to the model, including both context frames and current observations.

\paragraph{Experiments Details.}
Although our model can tolerate moderate camera pose variation, the two external cameras are expected to provide approximately front and side views of the workspace. Since RLBench does not include a side-view camera by default, we add an additional camera on the robot's right side, with its viewing direction approximately orthogonal to the front-view camera. We use the same cameras extrinsics for generating data in PyBullet. For real-world tasks, we place the cameras to roughly match the Isaac Sim camera setup, with one camera in front of the workspace and one camera on the robot's left side. However, we find that the model still maintains similar performance under additional slight perturbations to the camera poses.

\section{Training Hyperparameters}
\begin{table}[h]
\centering
\caption{Training hyperparameters for SynthICL.}
\begin{tabular}{l c}
\hline
\textbf{Hyperparameter} & \textbf{Value} \\
\hline
Image resolution & $256 \times 256$ \\
Prediction horizon & 8 \\
Action dimension & 7 \\
State dimension & 7 \\
Batch size & 64 \\
Optimizer & AdamW \\
Learning rate & $1 \times 10^{-4}$ \\
Gradient clipping & 1.0 \\
Mixed precision & yes \\
Training epochs & 2000 \\
Vision backbone & frozen DINOv3 (ViT-S/16+) \\
Transformer dimension & 512 \\
Context encoder layers & 8 \\
State-context encoder layers & 7 \\
Action decoder layers & 2 \\
Subgoal decoder layers & 2 \\
Attention heads & 8 \\
Action time sampling & Beta$(1.5, 1.0)$ \\
Time range & $[0.001, 0.999]$ \\
Auxiliary loss weight & 1.0 \\
Inference flow steps & 10 \\
\hline
\end{tabular}
\end{table}

\section{Real-World Task Descriptions}
\label{sec:real_world_task_descriptions}

We evaluate SynthICL on a set of real-world manipulation tasks that require visual grounding, object interaction, and precise end-effector control. Each task is specified by a single test-time demonstration, and the policy must infer the task goal from the provided visual context and execute the task from a new initial configuration. Below, we describe the details of each real-world task.

\paragraph{Mug on Stove.}
The robot is required to pick up a mug by its handle from the tabletop and place it onto the stove. The stove contains two burners, and the robot must place the mug on the target burner indicated by the demonstration. 

\paragraph{Bread Transfer.}
The robot is required to pick up a piece of bread from one toaster and insert it into another toaster. 

\paragraph{Insert Paper Roll.}
The robot is required to grasp a paper roll and insert it into a target paper holder. 

\paragraph{Ball in Drawer.}
The robot is required to pick up a ball and place it inside an open drawer. 

\paragraph{Stack Cups.}
The robot is required to stack the target cup (indicated by demonstration) onto another cup. The two cups have similar shapes but different textures.

\paragraph{Stack Bowls.}
The robot is required to stack the target bowl (indicated by demonstration) onto another bowl. Similar to the cup-stacking task, the two bowls have similar shapes but different textures.

\paragraph{Five Holes Placement.}
The robot is required to pick up a cube and insert it into the target hole out of five possible holes (indicated by demonstration).

\paragraph{Hang Cup.}
The robot is required to grasp a cup and hang it on a hook.

\paragraph{Pour oil on Plate.}
The robot is required to pick up a toy pan and pour the oil on the pan to the plate.

\paragraph{Press Button.}
The robot is required to move to a button and press it.

\paragraph{Open Cash Register.}
The robot is required to press the button on the cash register to open the cash drawer.

\paragraph{Scanner on Base.}
The robot is required to grasp the scanner and place it onto the scanner base on the cash register.

\section{Failure Case Analysis}
\paragraph{Case 1: Objects with Identical Geometry.}
Although our policy uses RGB observations, we find that it can still struggle to distinguish objects with identical geometry but different colors. One example is the \textit{Stack Cups} task in RLBench, where multiple cups share the same shape but differ in appearance. This may be because the frozen DINOv3 encoder emphasizes geometric and semantic structure more strongly than fine-grained color or texture cues.

\paragraph{Case 2: Limited View Coverage for Precise Manipulation.}
For tasks that require accurate placement, such as \textit{Cube in Shape Sorter}, the policy tends to fail more often when the target placement location is far from both external cameras. This may be because, when the target region is far from the cameras, it occupies fewer pixels and contains less detailed visual information. As a result, the extracted visual features provide weaker spatial cues for precise localization, making accurate placement more difficult. One possible solution is to add a third external camera to provide better coverage of the workspace.

\paragraph{Case 3: Collision Avoidance.}
Since our pseudo-demonstrations do not explicitly model physical contacts or collision-free path planning, the policy receives limited supervision for collision avoidance. This can lead to failures when the target object or placement location is close to obstacles, where the robot may collide with surrounding objects while moving toward the goal.

\section{Qualitative Subgoal Prediction Results}
\begin{figure}[htbp]
    \centering
    \includegraphics[width=\linewidth]{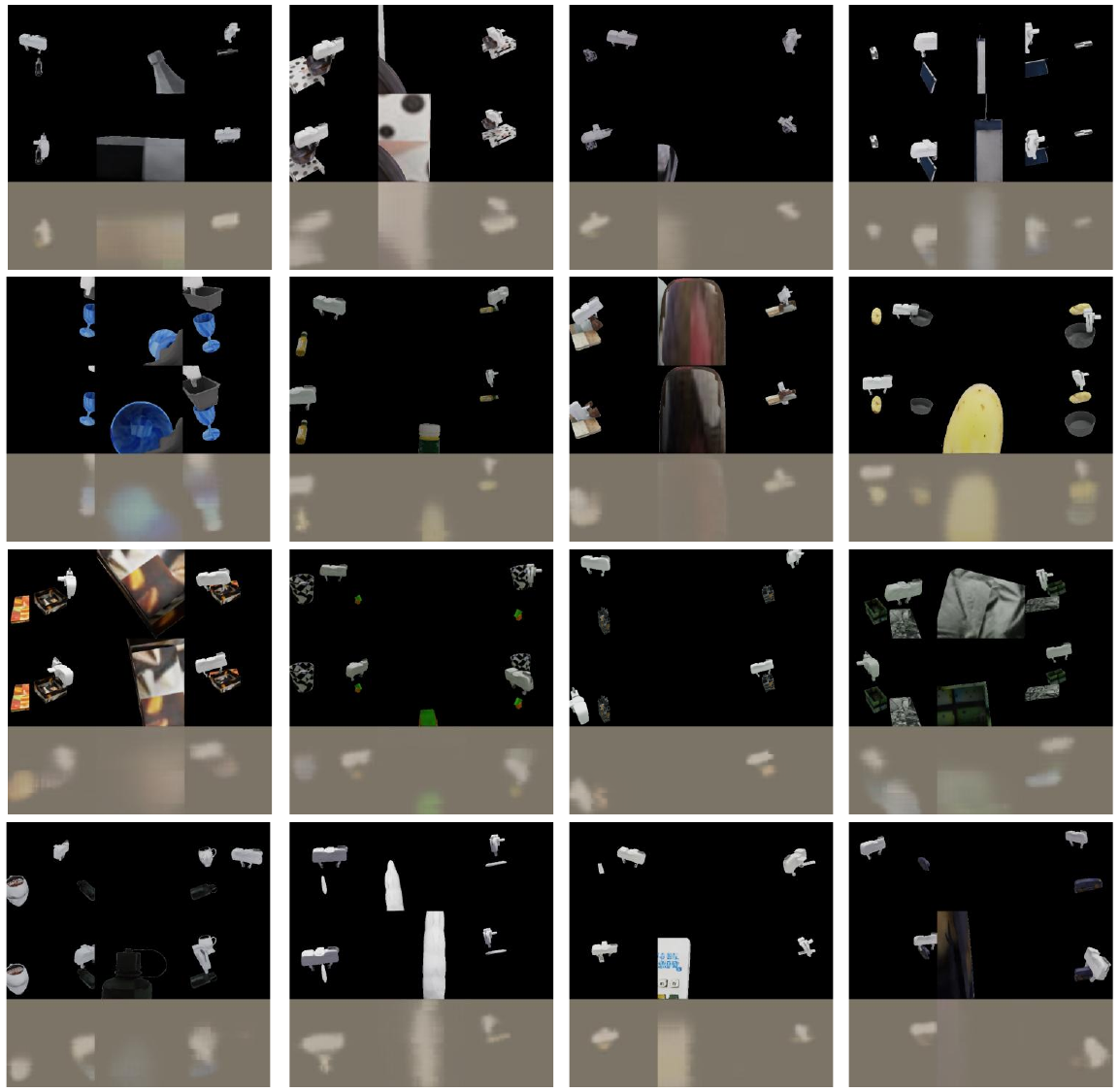}
    \caption{Each panel visualizes the predicted subgoal images. In each panel, the first row shows the multi-view current observation, the second row shows the ground-truth next subgoal images, and the third row shows the predicted next subgoal images.}
    \label{fig:subgoal_result}
\end{figure}

In Figure~\ref{fig:subgoal_result}, we visualize several subgoal prediction results from evaluation batches. The image backgrounds are black because they are segmented before being passed to the model. Although the predicted images are blurry, they still capture the overall object layout and task-relevant geometry. This blurriness may be caused by several factors. First, we train the subgoal prediction head with a simple MSE loss rather than a generative objective, e.g., diffusion or flow matching. Second, the prediction head reconstructs images from DINO features, which may emphasize geometric and semantic structure over fine-grained appearance details. Third, the subgoal prediction head is implemented as a lightweight transformer, which may limit its reconstruction capacity. However, importantly, the purpose of this auxiliary branch is not photorealistic image generation. Instead, it provides representation-level supervision that encourages the shared encoder to preserve information about the next visually meaningful task state.

\begin{table}[t]
\centering
\caption{Additional ablations in simulation. We report the average success rate across simulation tasks.}
\label{tab:additional_ablations}

\begin{subtable}{0.48\linewidth}
\centering
\caption{Model size}
\label{tab:ablation_model_size}
\begin{tabular}{l c}
\hline
\textbf{Model} & \textbf{Avg. Success} \\
\hline
Full model & 75.0\% \\
Smaller model & 64.8\% \\
\hline
\end{tabular}
\end{subtable}
\hfill
\begin{subtable}{0.48\linewidth}
\centering
\caption{DINOv3 tuning}
\label{tab:ablation_lora}
\begin{tabular}{l c}
\hline
\textbf{Variant} & \textbf{Avg. Success} \\
\hline
Frozen DINOv3 & 75.0\% \\
LoRA fine-tuning & 20.7\% \\
\hline
\end{tabular}
\end{subtable}

\vspace{0.8em}

\begin{subtable}{0.48\linewidth}
\centering
\caption{Execution strategy}
\label{tab:ablation_temporal_ensembling}
\begin{tabular}{l c}
\hline
\textbf{Variant} & \textbf{Avg. Success} \\
\hline
Temporal ensembling & 15/20 \\
Direct, $H=1$ & 12/20 \\
Direct, $H=2$ & 13/20 \\
Direct, $H=4$ & 13/20 \\
Direct, $H=8$ & 11/20 \\
\hline
\end{tabular}
\end{subtable}
\hfill
\begin{subtable}{0.48\linewidth}
\centering
\caption{Camera views}
\label{tab:ablation_camera_views}
\begin{tabular}{l c}
\hline
\textbf{Views} & \textbf{Avg. Success} \\
\hline
Front + side + wrist & 75.0\% \\
Front + side & 60.3\% \\
Front + wrist & 64.6\% \\
\hline
\end{tabular}
\end{subtable}

\end{table}

\section{Additional Ablations}
\paragraph{Smaller Model Size.}
We investigate a smaller model variant to reduce computation. Specifically, we reduce the number of context encoder layers from 8 to 6, the state-context encoder layers from 7 to 5, the action decoder layers from 2 to 1, and the latent dimension from 512 to 256. As shown in Table~\ref{tab:ablation_model_size}, reducing the model size decreases the average simulation success rate from 75.0\% to 64.8\%. We also observe qualitatively that the smaller model produces less accurate behavior than the full model. Interestingly, only the full model consistently learns recovery behaviors, such as regrasping.

\paragraph{LoRA Fine-Tuning of the DINOv3 Encoder.}
We experiment with LoRA fine-tuning of the DINOv3 encoder during policy training to improve visual adaptation. However, this substantially degrades performance: the average simulation success rate drops to 20.7\% with LoRA fine-tuning, as reported in Table~\ref{tab:ablation_lora}. One possible reason is that the same DINOv3 encoder is shared across all three camera views, while the visual distributions of these views differ substantially, especially for the wrist camera. As a result, a single shared LoRA adaptation may struggle to improve all views simultaneously and can degrade the general visual representation.

\paragraph{Executing Action Chunks without Temporal Ensembling.}
We also experiment with directly executing action chunks with horizons of 1, 2, 4, and 8, without temporal ensembling~\citep{zhao2023learning}. While these variants perform similarly in simulation, we find that direct execution is less robust in precision-required real-world tasks. We therefore evaluate this design choice on the real-world \textit{Cube in Shape Sorter} task. As shown in Table~\ref{tab:ablation_temporal_ensembling}, using temporal ensembling achieves the highest success rate. We observe shakier and less robust behavior without temporal ensembling, especially when the gripper approaches the placement location. Therefore, we use temporal ensembling in all main experiments.

\paragraph{Using Fewer Camera Views.}
We also evaluate variants that remove either the side camera or the wrist camera during both training and inference. As shown in Table~\ref{tab:ablation_camera_views}, removing the wrist camera reduces the average simulation success rate to 60.3\%, while removing the side camera reduces it to 64.6\%. These results suggest that the three views provide complementary information. We observe that removing the wrist camera makes it difficult for the robot to align precisely with the target object, especially during grasping, since the policy loses close-up visual feedback near the gripper. Removing the side camera mainly hurts depth perception and spatial localization, leading to inaccurate placement behavior.

\end{document}